\pdfoutput=1

\documentclass{article}

\newcommand{\argmin}{\mathop{\mathrm{argmin\,}}}

\newcommand{\boldC}{{\boldsymbol{C}}}
\newcommand{\boldD}{{\boldsymbol{D}}}

\newcommand{\boldI}{{\boldsymbol{I}}}

\newcommand{\boldM}{{\boldsymbol{M}}}

\newcommand{\boldP}{{\boldsymbol{P}}}

\newcommand{\boldR}{{\boldsymbol{R}}}

\newcommand{\boldU}{{\boldsymbol{U}}}

\newcommand{\boldX}{{\boldsymbol{X}}}
\newcommand{\boldY}{{\boldsymbol{Y}}}
\newcommand{\boldZ}{{\boldsymbol{Z}}}

\newcommand{\bolda}{{\boldsymbol{a}}}

\newcommand{\boldd}{{\boldsymbol{d}}}
\newcommand{\bolde}{{\boldsymbol{e}}}

\newcommand{\boldh}{{\boldsymbol{h}}}

\newcommand{\boldm}{{\boldsymbol{m}}}

\newcommand{\boldu}{{\boldsymbol{u}}}
\newcommand{\boldv}{{\boldsymbol{v}}}

\newcommand{\boldx}{{\boldsymbol{x}}}

\newcommand{\boldz}{{\boldsymbol{z}}}

\usepackage[square, numbers]{natbib}
\usepackage{microtype}
\usepackage{graphicx}
\usepackage{subfigure}
\usepackage{booktabs} 

\usepackage{hyperref}

\usepackage[accepted]{icml2023}

\usepackage{amsmath}
\usepackage{amssymb}
\usepackage{mathtools}
\usepackage{amsthm}
\usepackage{bbm}

\usepackage{tcolorbox}

\usepackage[capitalize,noabbrev]{cleveref}

\newcommand*{\lbb}{\{\mskip-5mu\{}
\newcommand*{\rbb}{\}\mskip-5mu\}}

\newenvironment{prooftn}[1]{%
\proof}{\endproof}

\newenvironment{sketch}{%
\proof}{\endproof}

\theoremstyle{plain}
\newtheorem{theorem}{Theorem}[section]

\newtheorem{lemma}[theorem]{Lemma}

\theoremstyle{definition}
\newtheorem{definition}[theorem]{Definition}

\theoremstyle{remark}

\usepackage[textsize=tiny]{todonotes}

\icmltitlerunning{GNNs can Recover the Hidden Features Solely from the Graph Structure}

\begin{document}

\twocolumn[
\icmltitle{Graph Neural Networks can Recover the Hidden Features \\ Solely from the Graph Structure}



\icmlsetsymbol{equal}{*}

\begin{icmlauthorlist}
\icmlauthor{Ryoma Sato}{kyoto,riken}
\end{icmlauthorlist}

\icmlaffiliation{kyoto}{Kyoto University}
\icmlaffiliation{riken}{RIKEN AIP}

\icmlcorrespondingauthor{Ryoma Sato}{r.sato@ml.ist.i.kyoto-u.ac.jp}

\icmlkeywords{Graph Neural Networks, Expressive Power, Metric Recovery}

\vskip 0.3in
]



\printAffiliationsAndNotice{\icmlEqualContribution} 

\begin{abstract}
Graph Neural Networks (GNNs) are popular models for graph learning problems. GNNs show strong empirical performance in many practical tasks. However, the theoretical properties have not been completely elucidated. In this paper, we investigate whether GNNs can exploit the graph structure from the perspective of the expressive power of GNNs. In our analysis, we consider graph generation processes that are controlled by hidden (or latent) node features, which contain all information about the graph structure. A typical example of this framework is kNN graphs constructed from the hidden features. In our main results, we show that GNNs can recover the hidden node features from the input graph alone, even when all node features, including the hidden features themselves and any indirect hints, are unavailable. GNNs can further use the recovered node features for downstream tasks. These results show that GNNs can fully exploit the graph structure by themselves, and in effect, GNNs can use both the hidden and explicit node features for downstream tasks. In the experiments, we confirm the validity of our results by showing that GNNs can accurately recover the hidden features using a GNN architecture built based on our theoretical analysis.
\end{abstract}

\section{Introduction}

\begin{figure*}[tb]
\centering
\includegraphics[width=\hsize]{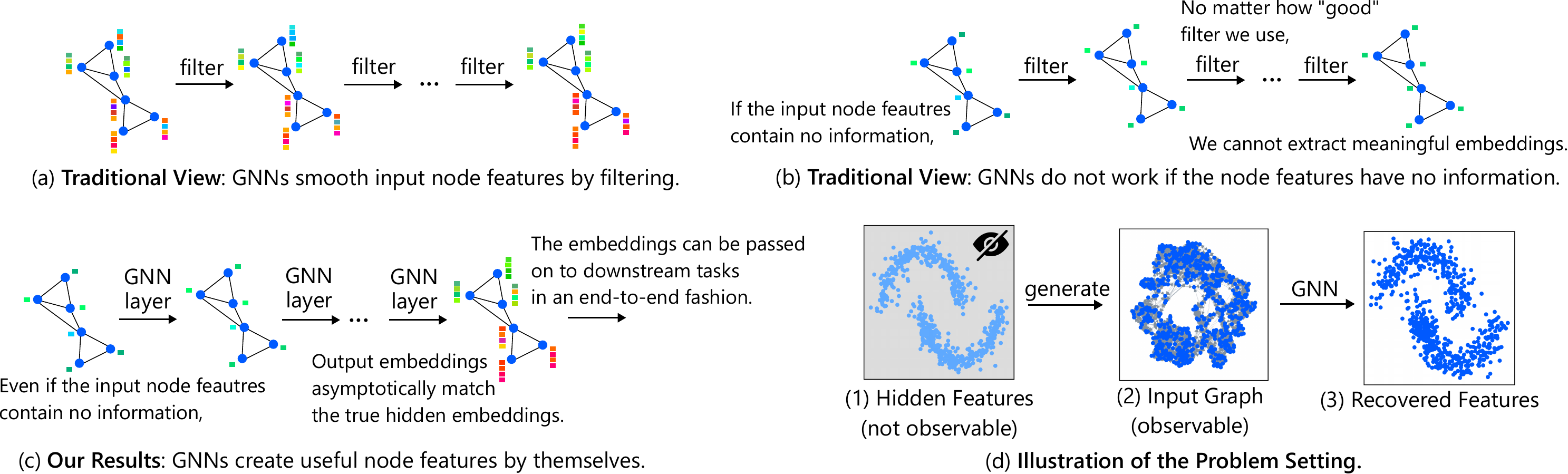}
\vspace{-0.2in}
\caption{(a) \textbf{Traditional View (Rich Node Features).} GNNs filter features by mixing them with neighboring nodes. (b) \textbf{Traditional View (Uninformative Features).} Filters cannot generate informative features if the inputs are not informative, i.e., garbage in, garbage out. (c) \textbf{Our Results.} GNNs create informative node features by themselves even when the input node features are uninformative by absorbing information from the underlying graph. (d) \textbf{Illustrations of the Problem Setting.} (d.1) Nodes have hidden features from which the input graph is generated. (d.2) The input to GNNs is a vanilla graph without any additional features. Nodes have coordinates for visualization in this panel, but these coordinates are neither fed to GNNs. (d.3) GNNs try to recover the hidden features. }
\label{fig: illustration}
\end{figure*}

Graph Neural Networks (GNNs) \cite{gori2005new, scarselli2009graph} are popular machine learning models for processing graph data. GNNs take a graph with node features as input and output embeddings of nodes. At each node, GNNs send the node features to neighboring nodes, aggregate the received features, and output the new node features \cite{gilmer2017neural}. In this way, GNNs produce valuable node embeddings that take neighboring nodes into account. GNNs show strong empirical performance in machine learning and data mining tasks \cite{zhang2018link, fan2019graph, hw2020light, han2022vision}.

Roughly speaking, GNNs smooth out the node features on the input graph by recursively mixing the node features of neighboring nodes, and GNNs thereby transform noisy features into clean ones (Figure \ref{fig: illustration} (a)). This smoothing effect has been observed empirically \cite{chen2020measuring, NT2019revisiting} and shown  theoretically \cite{li2018deeper, oono2020graph}. There are several GNN architectures that are inspired by the smoothing process \cite{klicpera2019predict, NT2019revisiting, wu2019simplifying}. It has also been pointed out that stacking too many layers harms the performance of GNNs due to the over-smoothing effect \cite{li2018deeper, simple2020chen}, which is caused by too much mixing of the node features.

In this perspective, node features are the primary actors in GNNs, and graphs are secondary. If node features are uninformative at all, GNNs should fail to obtain meaningful node embeddings no matter how they mix node features (Figure \ref{fig: illustration} (b)). This is in contrast to the opposite scenario: Even if the graphs are uninformative at all, if the node features are informative for downstream tasks, GNNs can obtain meaningful node embeddings just by ignoring edges or not mixing node features at all. Therefore, node features are the first requirement of GNNs, and the graph only provides some boost to the quality of node features \cite{NT2019revisiting}. It indicates that GNNs cannot utilize the graph information without the aid of good node features.

The central research question of this paper is as follows:
\begin{center}
    \textit{Can GNNs utilize the graph information \\ without the aid of node features?}
\end{center}
We positively answer this question through our theoretical analysis. We show that GNNs can recover the hidden node features that control the generation of the graph structure even without the help of informative node features (Figure \ref{fig: illustration} (c)). The recovered features contain all the information of the graph structure. The recovered node features can be further used for downstream tasks. These results show that GNNs can essentially use both given node features and graph-based features extracted from the graph structure. Our theoretical results provide a different perspective from the existing beliefs \cite{wang2020am, NT2019revisiting} based on empirical observations that GNNs only mix and smooth out node features. In the experiments, we show that existing GNN architectures do not necessarily extract the hidden node features well, and special architectures are required to learn the recovery in empirical situations.

The contributions of this paper are summarized as follows:
\begin{itemize}
\item We establish the theory of the feature recovery problem by GNNs for the first time. Our analysis provides a new perspective on the expressive power of GNNs.
\item We prove that GNNs can recover the hidden features solely from the graph structure (Theorem \ref{thm: main}). These results show that GNNs have an inherent ability to extract information from the input graph.
\item We validate the theoretical results in the experiments by showing that GNNs can accurately recover the hidden features. We also show that existing GNN architectures are mediocre in this task. These results highlight the importance of inductive biases for GNNs.
\end{itemize}

\begin{tcolorbox}[colframe=gray!20,colback=gray!20,sharp corners]
\textbf{Reproducibility}: Our code is publicly available at \url{https://github.com/joisino/gnnrecover}.
\end{tcolorbox}

\section{Related Work}

\subsection{Graph Neural Networks and Its Theory}

Graph Neural Networks (GNNs) \cite{gori2005new, scarselli2009graph} are now de facto standard models for graph learning problems \cite{kipf2017semi, velickovic2018graph, zhang2018link}. There are many applications of GNNs, including bioinformatics \cite{li2021structure}, physics \cite{cranmer2020discovering, pfaff2021learning}, recommender systems \cite{fan2019graph, hw2020light}, and transportation \cite{wang2020traffic}. There are several formulations of GNNs, including spectral \cite{defferrard2016convolutional}, spatial \cite{gilmer2017neural}, and equivariant \cite{maron2019invariant} ones. We use the message-passing formulation \cite{gilmer2017neural} in this paper.

The theory of GNNs has been studied extensively in the literature, including generalization GNNs \cite{scarselli2018vapnik, garg2020generalization, xu2020what} and computational complexity \cite{hamilton2017inductive, chen2018fastgcn, zou2019layer, sato2022constant}. The most relevant topic to this paper is the expressive power of GNNs, which we will review in the following.

\textbf{Expressive Power} (or Representation Power) means what kind of functional classes GNNs can realize. Originally, Morris et al. \cite{morris2019weisfeiler} and Xu et al. \cite{xu2019how} showed that message-passing GNNs are at most as powerful as the 1-WL test, and they proposed GNNs that are as powerful as the 1-WL and $k$-(set)WL tests. Sato \cite{sato2019approximation, sato2021random} and Loukas \cite{loukas2020what} also showed that message-passing GNNs are as powerful as a computational model of distributed local algorithms, and they proposed GNNs that are as powerful as port-numbering and randomized local algorithms. Loukas \cite{loukas2020what} showed that GNNs are Turing-complete under certain conditions (i.e., with unique node ids and infinitely increasing depths). There are various efforts to improve the expressive power of GNNs by non-message-passing architectures \cite{maron2019invariant, maron2019provably, murphy2019relational}. We refer the readers to survey papers \cite{sato2020survey, jegelka2022theory} for more details on the expressive power of GNNs.

The main difference between our analysis and existing ones is that the existing analyses focus on combinatorial characteristics of the expressive power of GNNs, e.g., the WL test, which are not necessarily aligned with the interests of realistic machine learning applications.
By contrast, we consider the continuous task of recovering the hidden features from the input graph, which is an important topic in machine learning in its own right \cite{tenenbaum2000global, belkin2003laplacian, sussman2014consistent, sato2022towards}. To the best of our knowledge, this is the first paper that reveals the expressive power of GNNs in the context of feature recovery. Furthermore, the existing analysis of expressive power does not take into account the complexity of the models. The existing analyses show that GNNs can solve certain problems, but they may be too complex to be learned by GNNs. By contrast, we show that the feature recovery problem can be solved with low complexity.

\subsection{Feature Recovery}

Estimation of hidden variables that control the generation process of data has been extensively studied in the machine learning literature \cite{tenenbaum2000global, sussman2014consistent, kingma2014auto}. These methods are sometimes used for dimensionality reduction, and the estimated features are fed to downstream models. In this paper, we consider the estimation of hidden embeddings from a graph observation \cite{alamgir2012shortest, luxburg2013density, terada2014local, hashimoto2015metric}. The critical difference between our analysis and the existing ones is that we investigate whether GNNs can represent a recovery algorithm, while the existing works propose general (non-GNN) algorithms that recover features. To the best of our knowledge, we are the first to establish the theory of feature recovery based on GNNs.

Many empirical works propose feature learning methods for GNNs \cite{hamilton2017inductive, velickovic2019deep, you2020when, hu2020gpt, qiu2020gcc}. The differences between these papers and ours are twofold. First, these methods are not proven to converge to the true features, while we consider a feature learning method that converges to the true features. Second, the existing methods rely heavily on the input node features while we do not assume any input node features. The latter point is important because how GNNs exploit the input graph structure is a central topic in the GNN literature, and sometimes GNNs are shown to NOT benefit from the graph structure \cite{errica2020fair}. By contrast, our results show that GNNs can extract meaningful information from the input graph from a different perspective than existing work.

\section{Background and Problem Formulation}

In this paper, we assume that each node $v$ has hidden (or latent) features $\boldz_v$, and the graph is generated by connecting nodes with similar hidden features. For example, (i) $\boldz_v \in \mathbb{R}^d$ represents the preference of person $v$ in social networks, (ii) $\boldz_v$ represents the topic of paper $v$ in citation networks, and (iii) $\boldz_v$ represents the geographic location of point $v$ in spatial networks.

The critical assumption of our problem setting is that the features $\{\boldz_v \mid v \in V\}$, such as the true preference of people and the true topic of papers, are not observed, but only the resulting graph $G$ is observed.

Somewhat surprisingly, we will show that GNNs that take the vanilla graph $G$ with only simple synthetic node features such as degree features $d_v$ and graph size $n = |V|$ can consistently estimate the hidden features $\{\boldz_v \mid v \in V\}$ (Figure \ref{fig: illustration} (d)).

In the following, we describe the assumptions on data and models in detail.

\subsection{Assumptions}

In this paper, we deal with directed graphs. Directed graphs are general, and undirected graphs can be converted to directed graphs by duplicating every edge in both directions. We assume that there is an arc from $v$ to $u$ if and only if $\|\boldz_{v} - \boldz_{u}\| < s(\boldz_{v})$ for a threshold function $s\colon \mathbb{R}^d \to \mathbb{R}$, i.e., nodes with similar hidden features are connected. It is also assumed that the hidden features $\{\boldz_v\}$ are sampled from an unknown distribution $p(\boldz)$ in an i.i.d. manner. As we consider the consistency of estimators or the behavior of estimators in a limit of infinite samples (nodes), we assume that a node $v_i$ and its features $\boldz_{v_i} \sim p(\boldz)$ are generated one by one, and we consider a series of graphs $G_1 = (V_1 = \{v_1\}, E_1), G_2 = (V_2 = \{v_1, v_2\}, E_2), \ldots, G_n = (V_n, E_n), \ldots$ with an increasing number of nodes. Formally, the data generation process and the assumptions are summarized as follows.

\begin{description}
\item[Assumption 1 (Domain)] The domain $\mathcal{Z}$ of the hidden features is a convex compact domain in $\mathbb{R}^d$ with smooth boundary $\partial \mathcal{Z}$.
\item[Assumption 2 (Graph Generation)] For each $i \in \mathbb{Z}_+$, $\boldz_{v_i}$ is sampled from $p(\boldv)$ in an i.i.d. manner. There is a directed edge from $v$ to $u$ in $G_n$ if and only if $\|\boldz_{v} - \boldz_{u}\| < s_n(\boldz_{v})$.
\item[Assumption 3 (Density)] The density $p(\boldz)$ is positive and differentiable with bounded $\nabla \log (p(\boldz))$ on $\mathcal{Z}$.
\item[Assumption 4 (Threshold Function)] There exists a deterministic continuous function $\bar{s}(x) > 0$ on $\bar{\mathcal{Z}}$ such that $g_n^{-1} s_n(x)$ converges uniformly to $\bar{s}(x)$ for some $g_n \in \mathbb{R}$ with $g_n \xrightarrow{n \to \infty} 0$ and $g_n n^{\frac{1}{n + 2}} \log^{- \frac{1}{d+2}} n \xrightarrow{n \to \infty} \infty$ almost surely.
\item[Assumption 5 (Stationary Distribution)] $n \pi_{G_n}(v)$ is uniformly equicontinuous almost surely, where $\pi_{G_n}(v)$ is the stationary distribution of random walks on $G_n$.
\end{description}

Note that these assumptions are common to \citep{hashimoto2015metric}. It should be noted that the threshold functions $s_n$ can be stochastic and/or dependent on the data as long as Assumption 4 holds. For example, $k$-NN graphs can be realized in this framework by setting $s(\boldz_{v})$ to be the distance to the $k$-th nearest neighbor from $\boldz_v$. We also note that Assumption 4 implies that the degree is the order of $\omega(n^{\frac{2}{d + 2}} \log^{\frac{d}{d+2}} n)$. Thus, the degree increases as the number of nodes increases. It ensures the graph is connected with high probability and is consistent with our scenario.

\textbf{Remark (One by One Generation).} The assumption of adding nodes one at a time may seem tricky. New users are indeed inserted into social networks one by one in some scenarios, but some other graphs do not necessarily follow this process. This assumption is introduced for technical convenience to consider the limit of $n \to \infty$ and to prove the consistency. In practice, the generation process of datasets does not need to follow this assumption. We use a single fixed graph in the experiments. GNNs succeed in recovering the hidden features only if the graph is sufficiently large.

\subsection{Graph Neural Networks}

We consider message-passing GNNs \cite{gilmer2017neural} in this paper. Formally, $L$-layer GNNs can be formulated as follows. Let $\boldX = [\boldx_1, \ldots, \boldx_n]^\top \in \mathbb{R}^{n \times d_{\text{in}}}$ be the explicit (i.e., given, observed) node features, and
\begin{align*}
    \boldh^{(0)}_v &= \boldx_v & (\forall v \in V), \notag \\
    \bolda^{(l)}_v &= f^{(l)}_{\text{agg}}(\lbb \boldh^{(l-1)}_u \mid u \in \mathcal{N}^-(v) \rbb) & (\forall l \in [L], v \in V), \\
    \boldh^{(l)}_v &= f^{(l)}_{\text{upd}}(\boldh^{(l-1)}_v, \bolda^{(l)}_v) & (\forall l \in [L], v \in V),
\end{align*}
where $\lbb \cdot \rbb$ denotes a multiset, and $\mathcal{N}^-(v)$ is the set of the neighbors with outgoing edges to node $v$. We call $f^{(l)}_{\text{agg}}$ an aggregation function and $f^{(l)}_{\text{upd}}$ a update function. Let $\theta = [L, f^{(1)}_{\text{agg}}, f^{(1)}_{\text{upd}}, \ldots, f^{(L)}_{\text{agg}}, f^{(L)}_{\text{upd}}]$ denote a list of all aggregation and update functions, i.e., $\theta$ specifies a model. Let $f(v, G, \boldX; \theta) = \boldh^{(L)}_v$ be the output of the GNN $\theta$ for node $v$ and input graph $G$. For notational convenience, $L_\theta$, $f^{(l)}_{\text{agg}, \theta}$, and $f^{(l)}_{\text{upd}, \theta}$ denote the number of layers, $l$-th aggregation function, and $l$-th aggregation function of model $\theta$, respectively.

Typical applications of GNNs assume that each node has rich explicit features $\boldx_v$. However, this is not the case in many applications, and only the graph structure $G$ is available. For example, when we analyze social networks, demographic features of users may be masked due to privacy concerns. In such a case, synthetic features that can be computed solely from the input graph, such as degree features and the number of nodes, are used as explicit node features $\boldx_v$ \cite{errica2020fair, hamilton2017inductive, xu2019how}. In this paper, we tackle this general and challenging setting to show how GNNs exploit the graph structure. Specifically, we do not assume any external node features but set
\begin{align} \label{eq: orig_feature}
\boldx_v = [d_v, n]^\top \in \mathbb{R}^2,
\end{align} where $d_v$ is the degree of node $v$, and $n = |V|$ is the number of nodes in $G$.

The goal of this paper is that GNNs can recover the hidden features $\boldz_v$ even if the node features are as scarce as the simple synthetic features. In words, we show that there exists GNN $\theta$ that uses the explicit node features $\boldX$ defined by Eq. \eqref{eq: orig_feature}\footnote{Precisely, we will add additional random features as Eq. \eqref{eq: extended_feature}.} and outputs $f(v, G, \boldX; \theta) \approx \boldz_v$. This result is surprising because GNNs have been considered to simply smooth out the input features along the input graph \cite{li2018deeper, NT2019revisiting}. Our results show that GNNs can imagine new features $\boldz_v$ that are not included in the explicit features $\boldX$ from scratch.

\textbf{Remark (Expressive Power and Optimization).} We note that the goal of this paper is to show the expressive power of GNNs, i.e., the existence of the parameters $\theta$ or the model specification that realizes some function, and how to find them from the data, i.e., optimization, is out of the scope of this paper. The separation of the studies of expressive power and optimization is a convention in the literature \cite{sato2019approximation, loukas2020what, abboud2021surprising}. This paper is in line with them. In the experiments, we briefly show the empirical results of optimization.

\subsection{Why Is Recovery Challenging?} \label{sec: challenging}

\begin{figure}[tb]
\centering
\includegraphics[width=0.85\hsize]{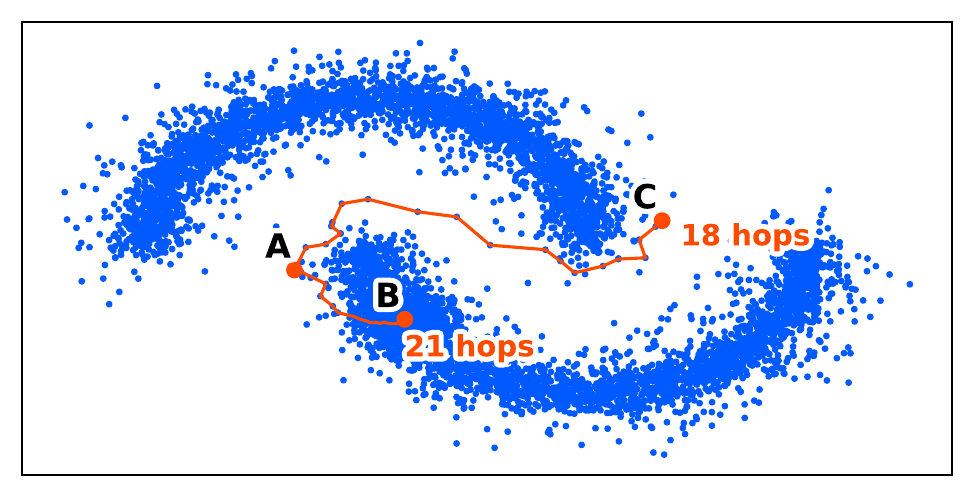}
\vspace{-0.15in}
\caption{\textbf{Illustrations of the Difficulty of Recovery.} The input graph is $10$-NN graph of the hidden features. The shortest path distance between points A and B is 21 hops, and the shortest path distance between points A and C is 18 hops. These distances indicate that point C is closer to point A than point B, but this is not the case in the true feature space. Standard node embedding methods would embed node C closer to A than node B to A, which is not consistent with the true feature. Embedding nodes that are close in the input graph close is the critical assumption in various embedding methods. This assumption does NOT hold in our situation. This disagreement is caused by the different scales of edges in sparse and dense regions. The difficulty lies in the fact that these scales are not directly available in the input information. }
\label{fig: shortest_path}
\end{figure}

The difficulty lies in the fact that the graph distance on the unweighted graph is not consistent with the distance in the hidden feature space. In other words, two nodes can be far in the hidden feature space even if they are close in the input graph. This fact is illustrated in Figure \ref{fig: shortest_path}. Most of the traditional node embedding methods rely on the assumption that close nodes in the graph should be embedded close. However, this assumption does not hold in our setting, and thus these methods fail to recover the hidden features from the vanilla graph. If the edge lengths $\|\boldz_v - \boldz_u\|$ in the hidden feature space are taken into account, the shortest-path distance on the graph is a consistent estimator for the distance of nodes in the hidden feature space. However, the problem is that the edge length $\|\boldz_v - \boldz_u\|$, such as the quantitative intimacy between people in social networks and the distance between the true topics of two papers in citation networks, is not available, and what we observe is only the vanilla graph $G$ in many applications. If we cannot rely on the distance structure in the given graph, it seems impossible to estimate the distance structure of the hidden feature space.

A hop count does not reflect the distance in the feature space because one hop on the graph in a sparse region is longer in the feature space than a hop in a dense region. The problem is, we do not know whether the region around node $v$ is dense because we do not know the hidden features $\boldz_v$. One might think that the density around a node could be estimated e.g., by the degree of the node, but this is not the case. Indeed, as the graph shown in Figure \ref{fig: shortest_path} is a $k$-NN graph, the degrees of all nodes are the same, and therefore the degree does not provide any information about the density. In general, the density cannot be estimated from a local structure, as von Luxburg et al. \cite{luxburg2013density} noted that ``It is impossible to estimate the density in an unweighted $k$-NN graph by local quantities alone.''

However, somewhat unexpectedly, it can be shown that the density function $p(\boldz)$ can be estimated solely from the unweighted graph \cite{hashimoto2015metric, stroock1971diffusion}. Intuitively, the random walk on the unweighted graph converges to the diffusion process on the true feature space $p(\boldz)$ as the number of nodes increases, and we can estimate the density function $p(\boldz)$ from it. Once the density is estimated, we can roughly estimate the threshold function $s(\boldz)$ as edges in low-density regions are long, and edges in dense regions are short. The scale function represents a typical scale of edges around $\boldz$ can be used as a surrogate value for the edge length.

Even if the scale function can be estimated in principle, it is another story whether GNNs can estimate it. We positively prove this. In the following, we first focus on estimating the threshold function $s(\boldz)$ by GNNs, and then, we show that GNNs can recover the hidden features $\boldz_v$ by leveraging the threshold function. 

\section{Main Results}

We present our main results and their proofs in this section. At a high level, our results are summarized as follows:
\begin{itemize}
    \item We show in Section \ref{sec: scale} that GNNs can estimate the threshold function $s(\boldz)$ with the aid of the metric recovery theory of unweighted graphs \cite{hashimoto2015metric, alamgir2012shortest}. We use the tool on the random walk and diffusion process, developed in \cite{hashimoto2015metric, stroock1971diffusion} 
    \item We show in Section \ref{sec: main} that GNNs can recover the hidden features up to rigid transformation with the aid of the theory of multidimensional scaling \cite{sibson1979studies, sibson1978studies} and random node features \cite{sato2021random, abboud2021surprising}.
    \item We show in Theorems \ref{thm: scale-bound} and \ref{thm: main-bound} that the number of the functions to be learned is finite regardless of the number of nodes, which is important for learning and generalization \cite{xu2020what}.
\end{itemize}

\subsection{Graph Neural Networks can Recover the Threshold Function} \label{sec: scale}

First, we show that GNNs can consistently estimate the threshold function $s$. As we mentioned in the previous section, the density and threshold function cannot be estimated solely from the local structure. As we will show (and as is known in the classical context), they can be estimated by a PageRank-like global quantity of the input graph.

\begin{theorem} \label{thm: scale}
    For any $s$ and $g$ that satisfy Assumptions 1-5, there exist $\theta_1, \theta_2, \ldots$ such that with the explicit node features $\boldX$ defined by Eq. \eqref{eq: orig_feature}, \begin{align*}
        \textup{Pr}\left[f(v, G_n, \boldX; \theta_n) \xrightarrow{n \to \infty} s(\boldz_v)\right] = 1,
    \end{align*} where the probability is with respect to the draw of samples $\boldz_1, \boldz_2, \ldots$.
\end{theorem}

\begin{sketch}
We prove this theorem by construction. The key idea is that GNNs can simulate random walks on graphs \cite{dehmamy2019understanding}. Once the stationary distribution of random walks is estimated, we can recover the scale from it \citep{hashimoto2015metric}. The full proof can be found in Appendix \ref{sec: proof_scale}.
\end{sketch}

This theorem states that GNNs can represent a consistent estimator of $s(\boldz_v)$.

However, this theorem does not bound the number of layers, and the number of layers may grow infinitely as the number of nodes increases. This means that if the size of the graphs is not bounded, the number of functions to be learned grows infinitely. This is undesirable for learning. The following theorem resolves this issue.

\begin{theorem} \label{thm: scale-bound}
    There exist $h^{(1)}_{\text{agg}}, h^{(1)}_{\text{upd}}, h^{(2)}_{\text{agg}}, h^{(2)}_{\text{upd}}$ such that for any $s$ and $g$, there exist $h^{(3)}_{\text{upd}}$ such that Theorem \ref{thm: scale} holds with \begin{alignat*}{3}
        f^{(1)}_{\text{agg}, \theta_n} = h^{(1)}_{\text{agg}}, & \quad & f^{(1)}_{\text{upd}, \theta_n} = h^{(1)}_{\text{upd}} & & \\
        f^{(l)}_{\text{agg}, \theta_n} = h^{(2)}_{\text{agg}}, & \quad & f^{(l)}_{\text{upd}, \theta_n} = h^{(2)}_{\text{upd}} & \quad & (l = 2, \ldots, L_{\theta_n} - 1) \\
        f^{(L_{\theta_n})}_{\text{agg}, \theta_n} = h^{(2)}_{\text{agg}}, & \quad & f^{(L_{\theta_n})}_{\text{upd}, \theta_n} = h^{(3)}_{\text{upd}}. & &
    \end{alignat*}
\end{theorem}

\begin{sketch}
From the proof of Theorem \ref{thm: scale}, most of the layers in $\theta_i$ are used for estimating the stationary distribution, which can be realized by a repetition of the same layer.
\end{sketch}

This theorem shows that the number of functions we need to learn is essentially five. This result indicates that learning the scale function has a good algorithmic alignment \citep[Definition 3.4]{xu2020what}. Moreover, these functions are the same regardless of the graph size. Therefore, in theory, one can fit these functions using small graphs and apply the resulting model to large graphs as long as the underlying law for the generation process, namely $s$ and $g$, is fixed. Note that the order of the logical quantification matters. As $h^{(1)}_{\text{agg}}, h^{(1)}_{\text{upd}}, h^{(2)}_{\text{agg}}, h^{(2)}_{\text{upd}}$ are universal and are independent with the generation process, they can be learned using other graphs and can be transferred to other types of graphs. The construction of these layers (i.e., the computation of the stationary distribution) can also be used for introducing indicative biases to GNN architectures. 

\subsection{Graph Neural Networks can Recover the Hidden Features} \label{sec: main}

As we have estimated the scale function, it seems easy to estimate the distance structure by applying the Bellman-Ford algorithm with edge lengths and to recover the hidden features, but this does not work well.

The first obstacle is that there is a freedom of rigid transformation. As rotating and shifting the true hidden features does not change the observed graph, we cannot distinguish hidden features that are transformed by rotation solely from the graph. To absorb the degree of freedom, we introduce the following measure of discrepancy of features.

\begin{definition}
We define the distance between two feature matrices $\boldX, \boldY \in \mathbb{R}^{n \times d}$ as \begin{align} \label{eq: d_G}
d_G(\boldX, \boldY) \stackrel{\text{def}}{=} \min_{\substack{\boldP \in \mathbb{R}^{d \times d} \\ \boldP^\top \boldP = I_d}} \frac{1}{n} \|\boldC_n \boldX - \boldC_n \boldY \boldP\|_F^2,
\end{align}
where $\boldC_n \stackrel{\text{def}}{=} (\boldI_n - \frac{1}{n}\mathbbm{1}_n \mathbbm{1}_n^\top) \in \mathbb{R}^{n \times n}$ is the centering matrix, $\boldI_n \in \mathbb{R}^{n \times n}$ is the identity matrix, and $\mathbbm{1}_n \in \mathbb{R}^n$ is the vector of ones. We say that we recover the hidden features $\boldX$ if we obtain features $\boldY$ such that $d_G(\boldX, \boldY) < \varepsilon$ for sufficiently small $\varepsilon > 0$. 
\end{definition}
In other words, the distance is the minimum average distance between two features after rigid transformation. This distance is sometimes referred to as the orthogonal Procrustes distance \cite{hurley1962procrustes, schonemann1966generalized, sibson1978studies}, and can be computed efficiently by SVD \cite{schonemann1966generalized}. Note that if one further wants to recover the rigid transformation factor, one can recover it in a semi-supervised manner by the Procrustes analysis.

The second obstacle is that GNNs cannot distinguish nodes. A naive solution is to include unique node ids in the node features. However, this leads the number of dimensions of node features to infinity as the number of nodes tends to infinity. This is not desirable for learning and generalization of the size of graphs. Our solution is to randomly select a constant number $m$ of nodes and assign unique node ids only to the selected nodes. Specifically, let $m \in \mathbb{Z}_+$ be a constant hyperparameter, and we first select $m$ nodes $\mathcal{U} = \{u_1, \ldots, u_m\} \subset V$ uniformly and randomly and set the input node features $\boldx_v \in \mathbb{R}^{2 + m}$ as \begin{align} \label{eq: extended_feature}
    \boldx_v^{\text{syn}} = \begin{cases}
        [d_v, n, \bolde_i^\top]^\top & (v = u_i) \\
        [d_v, n, \bold0_m^\top]^\top & (v \not \in \mathcal{U})
    \end{cases},
\end{align} where $\bolde_i \in \mathbb{R}^m$ is the $i$-th standard basis, and $\bold0_m$ is the vector of zeros. Importantly, this approach does not increase the number of dimensions even if the number of nodes tends to infinity because $m$ is a constant with respect to $n$. From a technical point of view, this is a critical difference from existing analyses \cite{loukas2020what, sato2021random, abboud2021surprising}, which assume unique node ids. Our analysis strikes an excellent trade-off between a small complexity (a constant dimension) and a strong expressive power (precise recovery). In addition, adding node ids have been though to be valid only for transductive settings \citep[Section 5.1.1]{hamilton2020graph}, but our analysis is valid for inductive setting as well (see also the experiments).

We show that we can accurately estimate the distance structure and the hidden features by setting an appropriate number of the selected nodes.

\begin{theorem} \label{thm: main}
    For any $s$ and $g$ that satisfy Assumptions 1-5, for any $\varepsilon, \delta > 0$, there exist $m$ and $\theta_1, \theta_2, \ldots$ such that with the explicit node features $\boldX$ defined by Eq. \eqref{eq: extended_feature}, \begin{align*}
        \textup{Pr}\left[\limsup_{n \to \infty} d_G(\hat{\boldZ}_{\theta_n}, \boldZ) < \varepsilon\right] > 1 - \delta,
    \end{align*} where $\hat{\boldZ}_{\theta_n} = [f(v_1, G_n, \boldX; \theta_n), \ldots, f(v_n, G_n, \boldX; \theta_n)]^\top \in \mathbb{R}^{n \times d}$ is the estimated hidden features by GNN $\theta_i$, and $\boldZ = [\boldz_1, \ldots, \boldz_n]^\top \in \mathbb{R}^{n \times d}$ is the true hidden features. The probability is with respect to the draw of samples $\boldz_1, \boldz_2, \ldots$ and the draw of a random selection of $\mathcal{U}$.
\end{theorem}

\begin{sketch}
We prove this theorem by construction. We estimate the threshold function $s$ by Theorem \ref{thm: scale} and compute the shortest path distances from each selected node in $\mathcal{U}$ with the estimated edge lengths. The computation of shortest path distances can be done by GNNs \cite{xu2020what}. After this process, each node has the information of the (approximate) distance matrix among the selected nodes, which consists of $m^2$ dimensions. We then run multidimensional scaling in each node independently and recover the coordinates of the selected nodes. Lastly, the selected nodes announce their coordinates, and the non-selected nodes output the coordinates of the closest nodes in $\mathcal{U}$. With sufficiently large $m$, the selected nodes $\mathcal{U}$ form an $\varepsilon'$-covering of $\mathcal{D}$ with high probability, and therefore, the mismatch of the non-selected nodes is negligibly small. The full proof can be found in Appendix \ref{sec: proof_main}.
\end{sketch}

As in Theorem \ref{thm: scale}, the statement of Theorem \ref{thm: main} does not bound the number of layers. However, as in Theorem \ref{thm: scale-bound}, Theorem \ref{thm: main} can also be realized with a fixed number of functions.

\begin{theorem} \label{thm: main-bound}
    For any $s$ and $g$, there exist $h^{(1)}_{\text{agg}}$, $h^{(1)}_{\text{upd}}$, $\ldots$, $h^{(8)}_{\text{agg}}$, $h^{(8)}_{\text{upd}}$ such that Theorem \ref{thm: main} holds with these functions.
\end{theorem}

Therefore, the number of functions we need to learn is essentially a constant.  This fact indicates that learning the hidden features has a good algorithmic alignment \citep[Definition 3.4]{xu2020what}. Besides, the components of these functions, i.e., computation of the stationary distribution, shortest-path distances, and multidimensional scaling, are differentiable almost everywhere. Here, we mean by almost everywhere the existence of non-differentiable points due to the min-operator of the shortest-path algorithm. Strictly speaking, this is no more differentiable than the ReLU function is, but can be optimized in an end-to-end manner by backpropagation using auto-differential frameworks such as PyTorch.

\begin{figure*}[tb]
\centering
\includegraphics[width=0.9\hsize]{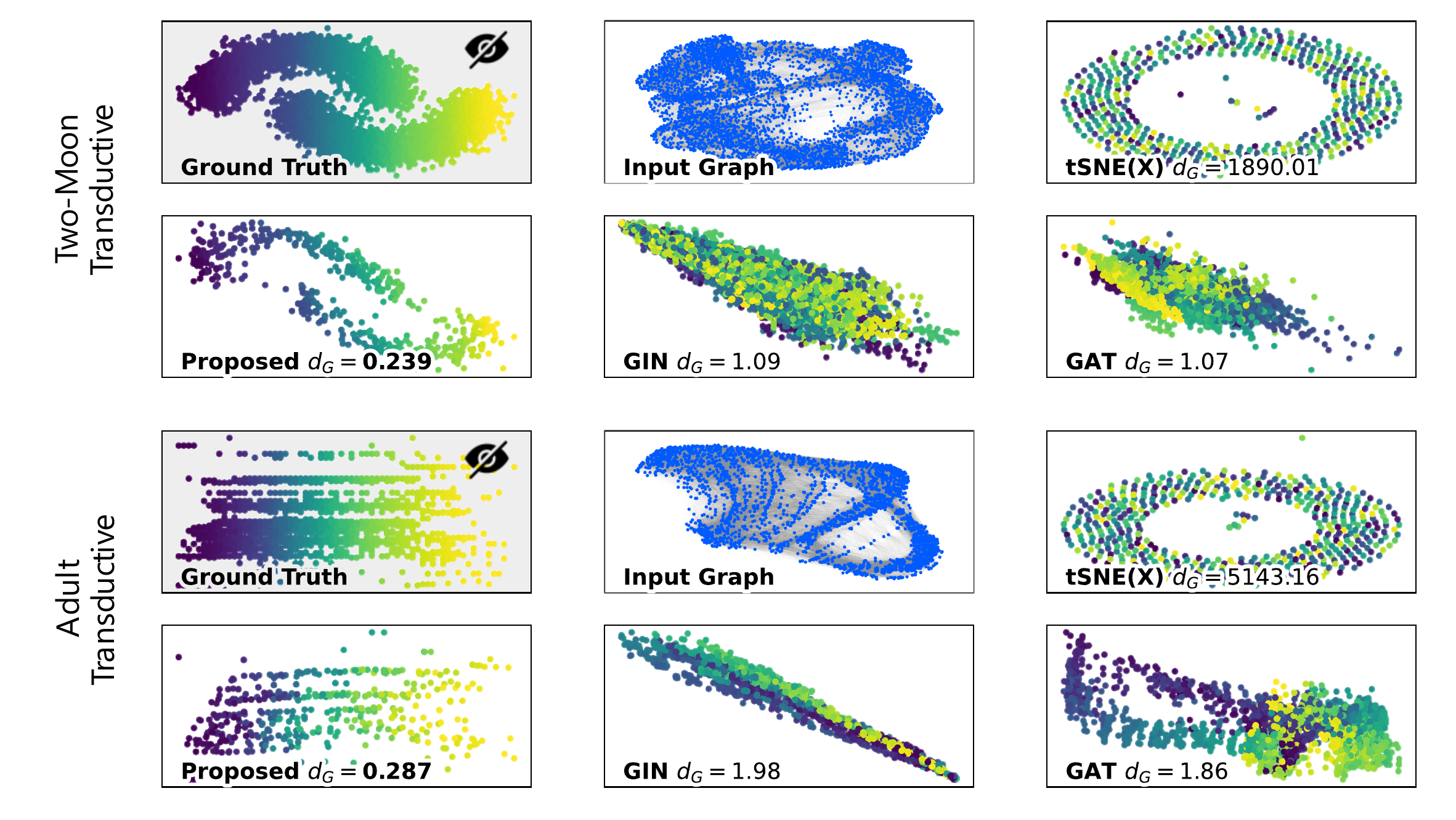}
\vspace{-0.1in}
\caption{\textbf{Results for the Transductive Setting.} Overall, the proposed method succeeded in recovering the ground truth hidden features while tSNE to $\boldX$ (Eq. \eqref{eq: extended_feature}) fails, and GINs and GATs are mediocre. (Top Left) The ground truth hidden embeddings. The node ids are numbered based on the x-coordinate and shown in the node colors. These node ids are for visualization purposes only and are NOT shown to GNNs and downstream algorithms. (Top Mid) The input graph constructed from the hidden features. The positions of the visualization are NOT shown to GNNs. (Top Right) tSNE plot on the synthetic node features, i.e., Eq. \eqref{eq: extended_feature}. These results indicate that the node features are not informative for feature recovery. This introduces challenges to the task. (Bottom Left) The recovered features by the proposed method. They resemble the ground truth not only with respect to the cluster structure but also the x-coordinates (shown in the node colors), the curved moon shapes in the two-moon dataset, and the striped pattern in the Adult dataset. The $d_G$ value (Eq. \eqref{eq: d_G}) is small, which indicates the success of the recovery and validates the theory. (Bottom Mid) The recovered features by GINs. They do not resemble the true hidden features. The $d_G$ value is mediocre. (Bottom Right) The recovered features by GATs. They do not resemble hidden features, but some clusters are detected (shown in the node colors). The $d_G$ value is mediocre. These results show that existing GNNs can extract some information from the graph structure, but they do not fully recover the hidden features. }
\vspace{-0.1in}
\label{fig: transductive}
\end{figure*}

\begin{figure*}[tb]
\centering
\includegraphics[width=0.9\hsize]{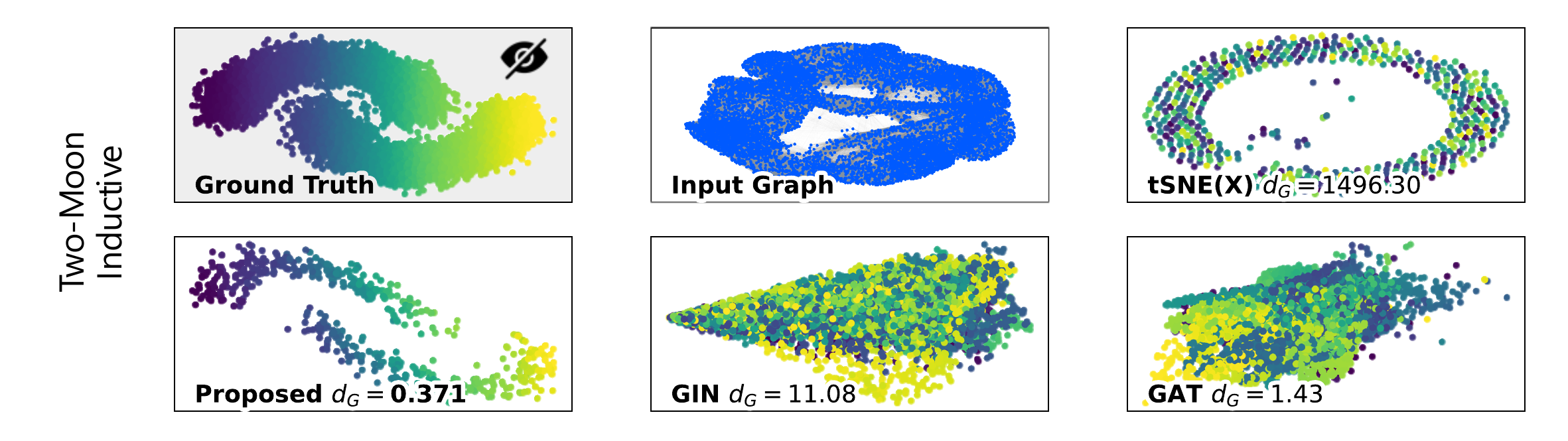}
\vspace{-0.2in}
\caption{\textbf{Results for the Inductive Setting.} The legends and tendencies are the same as in Figure \ref{fig: transductive}. The proposed method succeeded in generalizing to different sizes and keeping $d_G$ low even in the extrapolation setting. GINs and GAT partially succeeded in extracting some of graph information, but they are not perfect, and $d_G$ is moderately high. }
\vspace{-0.2in}
\label{fig: inductive}
\end{figure*}

\textbf{Remark (GNNs with Input Node Features).} Many of graph-related tasks provide node features $\{\boldx_v^{\text{given}} \mid v \in V\}$ as input. Theorem \ref{thm: main} shows that GNNs with $\boldx_v^{\text{syn}}$ as explicit node features can recover $\boldz_v$, where $\boldx_v^{\text{syn}}$ is defined by Eq. \eqref{eq: extended_feature}. Thus, if we feed $\boldx_v = [\boldx_v^{\text{given}\top}, \boldx_v^{\text{syn}\top}]^\top$ to GNNs as explicit node features, GNN can implicitly use both of $\boldx_v^{\text{given}}$ and $\boldz_v$.

The most straightforward method for node classification is to apply a feed-forward network to each node independently, \begin{align} \label{eq: mlp-classifier}
    \hat{y}_v^{\text{MLP}} = h_{\theta}(\boldx_v^{\text{given}}).
\end{align} This approach is fairly strong when $\boldx_v^{\text{given}}$ is rich \cite{NT2019revisiting} but ignores the graph. Our analysis shows that GNNs can classify nodes using both $\boldx_v^{\text{given}}$ and $\boldz_v$, i.e., \begin{align} \label{eq: gnn-classifier}
    \hat{y}_v^{\text{GNN}} = h_{\theta}(\boldx_v^{\text{given}}, \boldz_v).
\end{align} Comparison of Eqs. \eqref{eq: mlp-classifier} and \eqref{eq: gnn-classifier} highlights a strength of GNNs compared to feed-forward networks.

\section{Experiments}
In the experiments, we validate the theorems by empirically showing that GNNs can recover hidden features solely from the input graph.

\subsection{Recovering Features}

\textbf{Datasets.} We use the following synthetic and real datasets.
\begin{description}
\item[Two-Moon] is a synthetic dataset with a two-moon shape. We construct a $k$-nearest neighbor graph with $k = \text{floor}(\frac{1}{10} n^{1/2} \log n)$, which satisfies Assumption 4. As we know the ground-truth generation process and can generate different graphs with the same law of data generation, we use this dataset for validating the theorems and showing generalization ability in an inductive setting.
\item[Adult] is a real consensus dataset. We use age and the logarithm of capital gain as the hidden features and construct a $k$-nearest neighbor graph, i.e., people with similar ages and incomes become friends.
\end{description}

\textbf{Settings.} We use two different problem settings.
\begin{description}
\item[Transductive] setting uses a single graph. We are given the true hidden features for some training nodes, and estimate the hidden features of other nodes. We use $70$ percent of the nodes for training and the rest of the nodes for testing.
\item[Inductive] setting uses multiple graphs. In the training phase, we are given two-moon datasets with $n = 1000$ to $5000$ nodes and their true hidden features. In the test phase, we are given a new two-moon dataset with $n = 10000$ nodes and estimate the hidden features of the test graph. This setting is challenging because (i) we do not know any hidden features of the test graphs, and (ii) models need to generalize to extrapolation in the size of the input graphs.
\end{description}

\textbf{Methods.} As we prove the theorems by construction and know the configuration of GNNs that recover the hidden features except for the unknown parameters about the ground truth data (i.e., the scale $g_n$ and the constant $c$ that depends on $p$ and $\bar{s}$), we use the model architecture that we constructed in our proof and model the unknown parameters, i.e., scaling factor $g_n$, using $3$-layer perceptron with hidden $128$ neurons that takes $n$ as input and output $g_n$. This model can be regarded as the GNNs with the maximum inductive bias for recovering the hidden features. We fix the number of the selected nodes $m = 500$ throughout the experiments.

\textbf{Baselines.} We use $3$-layer Graph Attention Networks (GATs) \cite{velickovic2018graph} and Graph Isomorphism Networks (GINs) \cite{xu2019how} as baselines. We feed the same explicit node features as in our method, i.e., Eq. \eqref{eq: extended_feature}, and use the hidden features as the target of the regression.

\textbf{Details.} We optimize all the methods with Adam \cite{kingma2015adam} with a learning rate of $0.001$ for $100$ epochs. The loss function is $d_G(\hat{\boldZ}_{\theta}[\text{train-mask}], \boldZ[\text{train-mask}])$, where $\hat{\boldZ}_{\theta} \in \mathbb{R}^{n \times d}$ is the output of GNNs, $\boldZ$ is the ground truth hidden embeddings, and train-mask extracts the coordinates of the training nodes.

\begin{table*}[tb]
    \centering
    \caption{\textbf{Performance on Downstream Tasks.} Each value represents accuracy. Higher is Better. The performance of the baseline method shows that $\boldx_v^{\text{syn}}$ does not contain any information for solving the downstream tasks. GNNs take such uninformative node features only as node features. Nevertheless, the recovered feature is highly predictive. This indicates that GNNs create completely new and useful node features by themselves even when the input node features are uninformative. }
    \begin{tabular}{lccccccc} \toprule
        & Cora & CiteSeer & PubMed & Coauthor CS & Coauthor Physics & Computers & Photo \\ \midrule
        Baseline $\boldx_v^{\text{syn}}$ & 0.122 & 0.231 & 0.355 & 0.066 & 0.307 & 0.185 & 0.207 \\
        Recovered Feature $\hat{\boldz}_v$ & \textbf{0.671} & \textbf{0.640} & \textbf{0.653} & \textbf{0.492} & \textbf{0.745} & \textbf{0.528} & \textbf{0.566} \\
        \bottomrule
    \end{tabular}
    \label{tab: downstream}
    \vspace{-0.2in}
\end{table*}

\textbf{Results.} Figures \ref{fig: transductive} and \ref{fig: inductive} show the results. As the rigid transformation factor cannot be determined, we align the recovered features using the orthogonal Procrustes analysis in the postprocessing. We make the following observations.

\textbf{Observation 1. Recovery Succeeded.} As the lower left panels show, the proposed method succeeds in recovering the hidden features solely from the input graphs. Notably, not only coarse structures such as connected components are recovered but also details such as the curved moon shape in the two-moon dataset and the striped pattern in the Adult dataset are recovered.

\textbf{Observation 2. Existing GNNs are mediocre.} As the lower right panels show, GINs and GATs extract some information from the graph structure, e.g., they map nearby nodes to similar embeddings as shown by the node colors, but they fail to recover the hidden features accurately regardless of their strong expressive powers. This is primarily because the input node features contain little information, which makes recovery difficult. These results highligt the importance of inductive biases of GNNs to exploit the hidden features.

\textbf{Observation 3. tSNE fails.} The upper right panels show that tSNE on the explicit node features $\boldX$ failed to extract meaningful structures. These results indicate that the synthetic node features (Eq. \eqref{eq: extended_feature}) do not tell anything about the hidden features, and GNNs recover the hidden features solely from the graph structure.

\textbf{Observation 4. Recovery Succeeded in the Inductive Setting.} Figure \ref{fig: inductive} shows that the proposed method succeeded in the inductive setting as well. This shows that the ability of recovery can be transferred to other graph sizes as long as the law of data generation is the same.

\subsection{Performance on Downstream Tasks}

We confirm that the recovered feature $\hat{\boldz}_v$ is useful for downstream tasks using popular benchmarks.

We use the Planetoid datasets (Cora, CiteSeer, PubMed) \cite{yang2016revisiting}, Coauthor datasets, and Amazon datasets \cite{schur2018pitfalls}. First, we discard all the node features in the datasets (e.g., the text information of the citation network). We then feed the vanilla graph to GNNs in the proposed way and recover the hidden node features by GNNs. We fit a logistic regression that estimates the node label $y_v$ from the recovered features $\hat{\boldz}_v$. As a baseline, we fit a logistic regression that estimates the node label $y_v$ from the input node feature $\boldx_v^{\text{syn}}$, i.e., Eq. \eqref{eq: extended_feature}. We use the standard train/val/test splits of these datasets, i.e., 20 training nodes per class. The accuracy in the test sets is shown in Table \ref{tab: downstream}.

These results show that the recovered features by GNNs are informative for downstream tasks while the input node features are not at all. This indicates that GNNs extract meaningful information solely from the graph structure. We stress that this problem setting where no node features are available is extremely challenging for GNNs. Recall that existing GNNs use the node features (e.g., the text information of the citation network) contained in these datasets, which we intentionally discard and do not use. The results above show that GNNs work well (somewhat unexpectedly) in such a challenging situation.

\section{Conclusion}
In this paper, we showed that GNNs can recover the hidden node features, which contain all information about the graph structure, solely from the graph input. These results provide a different perspective from the existing results, which indicate that GNNs simply mix and smooth out the given node features. In the experiments, GNNs accurately recover the hidden features in both transductive and inductive settings.

\textbf{Acknowledgements.} This work was supported by JSPS KAKENHI GrantNumber 21J22490 and 20H04244. 

\bibliography{example_paper}

\newpage
\appendix
\onecolumn
\section{Proof of Theorems \ref{thm: scale} and \ref{thm: scale-bound}} \label{sec: proof_scale}

First, we introduce the following lemma.

\begin{lemma}{\cite{hashimoto2015metric}}
Under assumptions 1-5, \begin{align} \label{eq: hashimoto_convergence}
\left(\frac{c d_v}{n^2 g_n^d \pi_{G_n, v}}\right)^{\frac{1}{d+2}} \to \bar{s}(\boldz_v)
\end{align}
holds almost surely, where $c \in \mathbb{R}$ is a constant that depends on $p$ and $\bar{s}$, and $\pi_{G_n, v}$ is the stationary distribution of random walks on $G_n$.
\end{lemma}

We then prove Theorem \ref{thm: scale}.

\begin{prooftn}{\ref{thm: scale}}
We prove the theorem by construction. Let \begin{align} \label{eq: scale_num_layer}
    L_n \stackrel{\text{def}}{=} \max_{G = (V, E), |V| = n} ~\min \left\{l \in \mathbb{Z}_+ ~\middle\vert~ \left\|\boldR_G^l \mathbbm{1}_n - n \pi_{G_n}\right\| \le \frac{1}{n}\right\},
\end{align} where $\boldR_G \in \mathbb{R}^{n \times n}$ is the random walk matrix of graph $G$. As $\boldR_G^l \mathbbm{1}_n \xrightarrow{l \to \infty} n \pi_{G_n}$, the set in the minimum is not empty. As $\{G = (V, E) \mid |V| = n\}$ is finite for every $n$, the outer max exists, and therefore $L_n$ exists for every $n$. We set $L_{\theta_n} = L_n$. In the following, we build a GNN whose embeddings of $l$-th layer $(1 \le l \le L_n - 1)$ is \begin{align} \label{eq: scale_GNN_invariant}
    \boldh_v^{(l)} = [d_v, n, (\boldR_{G_n}^l \mathbbm{1}_n)_v]^\top.
\end{align}
The aggregation function of the first layer is \begin{align} \label{eq: scale_first_agg}
    f_{\text{agg}, \theta_n}^{(1)}\left(\lbb [d_u, n] \mid u \in \mathcal{N}^-(v) \rbb\right) \stackrel{\text{def}}{=} \sum_{u \in \mathcal{N}^-(v)} \frac{1}{d_u},
\end{align} i.e., $f_{\text{agg}, \theta_n}^{(1)}$ computes $(\boldR_{G_n} \mathbbm{1}_n)_v$, which is the sum of probabilities from the incoming edges. The update function of the first layer is \begin{align} \label{eq: scale_first_upd}
    f_{\text{upd}, \theta_n}^{(1)}\left([d_v, n]^\top, \bolda_v^{(1)}\right) \stackrel{\text{def}}{=} [d_v, n, \bolda_v^{(1)}]^\top.
\end{align} $\boldh^{(1)}_v$ holds condition \eqref{eq: scale_GNN_invariant} by construction. The aggregation function of the $l$-th layer ($2 \le l \le L_n$) is \begin{align} \label{eq: scale_mid_agg}
    f_{\text{agg}, \theta_n}^{(l)}\left(\lbb [d_u, n, (\boldR_{G_n}^{l-1} \mathbbm{1}_n)_u] \mid u \in \mathcal{N}^-(v) \rbb\right) = \sum_{u \in \mathcal{N}^-(v)} \frac{(\boldR_{G_n}^{l-1} \mathbbm{1}_n)_u}{d_u} \stackrel{\text{def}}{=} (\boldR_{G_n}^{l} \mathbbm{1}_n)_v,
\end{align} and the update function of the $l$-th layer ($2 \le l \le L_n - 1$) is \begin{align} \label{eq: scale_mid_upd}
    f_{\text{upd}, \theta_n}^{(l)}([d_v, n, (\boldR_{G_n}^{l-1} \mathbbm{1}_n)_v]^\top, \bolda_v^{(l)}) \stackrel{\text{def}}{=} [d_v, n, \bolda_v^{(l)}]^\top.
\end{align} $\boldh^{(l)}_v ~(2 \le l \le L_n - 1)$ holds condition \eqref{eq: scale_GNN_invariant} by construction. Lastly, the aggregation function of the $L_n$-th layer is \begin{align} \label{eq: scale_last_layer}
    f_{\text{upd}, \theta_n}^{(L_n)}\left(\left[d_v, n, (\boldR_{G_n}^{l-1} \mathbbm{1}_n)_v\right]^\top, \bolda_v^{(L_n)}\right) \stackrel{\text{def}}{=} \left(\frac{c d_v}{n g_n^d \bolda_v^{(l)}}\right)^{\frac{1}{d+2}}.
\end{align} By Eq. \eqref{eq: scale_num_layer} and \eqref{eq: scale_GNN_invariant}, $|\bolda_v^{(L_n)} - n \pi_{G_n}| \xrightarrow{n \to \infty} 0$ surely. Combining with Eq. \eqref{eq: hashimoto_convergence} and \eqref{eq: scale_last_layer} yields $f(v, G_n, \boldX; \theta_n) = \boldh^{(L_n)}_v \xrightarrow{n \to \infty} \bar{s}(\boldz_v)$ almost surely.
\end{prooftn}

The definitions of the layers, i.e., equations \eqref{eq: scale_first_agg}, \eqref{eq: scale_first_upd}, \eqref{eq: scale_mid_agg}, \eqref{eq: scale_mid_upd}, and \eqref{eq: scale_last_layer}, prove Theorem \ref{thm: scale-bound}.

\section{Proof of Theorems \ref{thm: main} and \ref{thm: main-bound}} \label{sec: proof_main}

\begin{prooftn}{\ref{thm: main}}
We prove the theorem by construction. Let $\mathcal{C}$ be an arbitrary $\frac{\varepsilon}{6}$ covering of $\mathcal{D}$. For each point $c \in \mathcal{C}$, let $\mathcal{B}(c; \frac{\varepsilon}{6}) \subset \mathbb{R}^d$ be the ball centered at $c$ with radius $\frac{\varepsilon}{6}$. The number $M_c$ of the selected points $\mathcal{U}$ in $\mathcal{B}(c; \frac{\varepsilon}{6})$ follows the binomial distribution $\text{Bi}(q_c, m)$, where $$q_c \stackrel{\text{def}}{=} \int_{\mathcal{B}(c; \frac{\varepsilon}{6})} p(x) dx$$ is positive. Therefore, $\text{Pr}[M_c = 0] = (1 - q_c)^m$. Let $$m \stackrel{\text{def}}{=} \text{ceil}\left(\frac{\log \frac{\delta}{2 |\mathcal{C}|}}{\log (1 - \min_{c \in \mathcal{C}} q_c)}\right),$$ then $$(1 - q_c)^m \le \frac{\delta}{2 |\mathcal{C}|},$$ and \begin{align*}
    \text{Pr}[\exists c \in \mathcal{C}, M_c = 0] \le |\mathcal{C}| (1 - q_c)^m \le \frac{\delta}{2}
\end{align*}
by the union bound. Therefore,
\begin{align*} 
    \text{Pr}[\forall c \in \mathcal{C}, M_c \ge 1] = 1 - \text{Pr}[\exists c \in \mathcal{C}, M_c = 0] \ge 1 - \frac{\delta}{2}.
\end{align*}
In words, with at least probability $1 - \frac{\delta}{2}$, each of $\mathcal{B}(c; \frac{\varepsilon}{6})$ contains at least one point. As $\mathcal{C}$ is an $\frac{\varepsilon}{6}$ covering of $\mathcal{D}$, $\mathcal{U}$ forms an $\frac{\varepsilon}{3}$ covering of $\mathcal{D}$ under $\forall c \in \mathcal{C}, M_c \ge 1$, i.e, 
\begin{align} \label{eq: prob_U_is_covering}
    \text{Pr}\left[\mathcal{U} \text{ forms an } \frac{\varepsilon}{3} \text{ covering}\right] \ge 1 - \frac{\delta}{2}.
\end{align}

We set $L_{\theta_n} = L_n + 2n$. The first $L_n$ layers are almost the same as the construction in Theorem \ref{thm: scale}. The only difference is that as we take $\boldx^{\text{id}}_v = \begin{cases} \bolde_i & (v = u_i) \\ \bold0_m & (v \not \in \mathcal{U}) \end{cases}$ as input (Eq. \eqref{eq: extended_feature}), we retain this information in the update functions. Therefore, in the $L_n$-th layer, each node has $\boldh_v^{(L_n)} = [\hat{\bar{s}}(\boldz_v), \boldx^{\text{id}\top}_v]^\top$ in the embedding, where, $\hat{\bar{s}}(\boldz_v)$ is the estimate of $\bar{s}(\boldz_v)$ computed by the GNN (Eq. \eqref{eq: scale_last_layer}).

The next $n$ layers compute the shortest-path distances from each node in $\mathcal{U}$ using the Bellman-Ford algorithm. Specifically, in the $L_n + 1$-th layer, the aggregation function is \begin{align} \label{eq: main_first_first_agg}
    f_{\text{agg}, \theta_n}^{(L_n + 1)}\left(\lbb [\hat{\bar{s}}(\boldz_u), \boldx^{\text{id}\top}_u]^\top \mid u \in \mathcal{N}^-(v) \rbb\right) &\stackrel{\text{def}}{=} \min_{u \in \mathcal{N}^-(v)} \boldm_u \in \mathbb{R}^m, \\
    \boldm_u &\stackrel{\text{def}}{=} \begin{cases} \left[\text{INF}, \ldots, \text{INF}, \overbrace{g_n \hat{\bar{s}}(\boldz_u)}^{i\text{th}}, \text{INF}, \ldots, \text{INF}\right] & \left(\boldx^{\text{id}}_{ui} = 1\right) \\
        [\text{INF}, \ldots, \text{INF}, \ldots, \text{INF}] & \left(\boldx^{\text{id}}_{u} = \bold0_m\right)
    \end{cases},
\end{align}
where min is element-wise minimum, and INF is a sufficiently large constant such as $\text{diam}(\mathcal{D}) + 1$. The update function of the $L_n + 1$-th layer is \begin{align} \label{eq: main_first_first_upd}
    f_{\text{upd}, \theta_n}^{(L_n + 1)}\left(\left[\hat{\bar{s}}(\boldz_v), \boldx^{\text{id}\top}_v\right]^\top, \bolda_v^{(L_n + 1)}\right) &\stackrel{\text{def}}{=} \left[\hat{\bar{s}}(\boldz_v), \boldx^{\text{id}\top}_v, \boldd_v^{(L_n + 1)\top}\right]^\top \in \mathbb{R}^{2m + 1}, \\
    \boldd_v^{(L_n + 1)} &\stackrel{\text{def}}{=} \min (\boldu_v, \bolda_v^{(L_n + 1)}), \\
    \boldu_v &\stackrel{\text{def}}{=} \begin{cases} \left[\text{INF}, \ldots, \text{INF}, \overbrace{0}^{i\text{th}}, \text{INF}, \ldots, \text{INF}\right] & \left(\boldx^{\text{id}}_{vi} = 1\right) \\
        \left[\text{INF}, \ldots, \text{INF}, \ldots, \text{INF}\right] & \left(\boldx^{\text{id}}_{u} = \bold0_m\right)
    \end{cases}.
\end{align}
The aggregation function of the $L_n + i$-th layer $(2 \le i \le n)$ is \begin{align} \label{eq: main_first_mid_agg}
    f_{\text{agg}, \theta_n}^{(L_n + i)}\left(\lbb [\hat{\bar{s}}(\boldz_u), \boldx^{\text{id}\top}_u, \boldd_u^{(L_n + i - 1)\top}]^\top \mid u \in \mathcal{N}^-(v) \rbb\right) &\stackrel{\text{def}}{=} \min_{u \in \mathcal{N}^-(v)} \boldd_u^{(L_n + i - 1)} + g_n \hat{\bar{s}}(\boldz_u),
\end{align}
and the update function is \begin{align} \label{eq: main_first_mid_upd}
    f_{\text{upd}, \theta_n}^{(L_n + i)}\left(\left[\hat{\bar{s}}(\boldz_v), \boldx^{\text{id}\top}_v, \boldd_v^{(L_n + i - 1)\top}\right]^\top, \bolda_v^{(L_n + i)}\right) &\stackrel{\text{def}}{=} \left[\hat{\bar{s}}(\boldz_v), \boldx^{\text{id}\top}_v, \boldd_v^{(L_n + i)\top}\right]^\top \in \mathbb{R}^{2m + 1}, \\
    \boldd_v^{(L_n + i)} &\stackrel{\text{def}}{=} \min \left(\boldd_v^{(L_n + i - 1)}, \bolda_v^{(L_n + i)}\right).
\end{align}
As the diameter of $G_n$ is at most $n$, the computation of the shortest path distance is complete after $n$ iterations. Therefore, ${\boldd_v^{(L_n + n)}}_j$ is the shortest-path distance from $u_j$ to $v$ with the length of edge $(s, t)$ being $g_n \hat{\bar{s}}(\boldz_s)$.

The following $n$ layers propagate the distance matrices among $\mathcal{U}$. Specifically, the aggregation function of the $L_n + n + 1$-th layer is \begin{align} \label{eq: main_mid_first_agg}
    &f_{\text{agg}, \theta_n}^{(L_n + 1)}\left(\lbb [\hat{\bar{s}}(\boldz_u), \boldx^{\text{id}\top}_u, \boldd_u^{(L_n + n)\top}]^\top \mid u \in \mathcal{N}^-(v) \rbb\right) \stackrel{\text{def}}{=} \min_{u \in \mathcal{N}^-(v)} \boldM_u \in \mathbb{R}^{m^2}, \\
    &\boldM_u \stackrel{\text{def}}{=} \begin{cases} \left[\text{INF} \mathbbm{1}_m^\top, \ldots, \text{INF} \mathbbm{1}_m^\top, \boldd_u^{(L_n + n)\top}, \text{INF}, \ldots, \text{INF}\right] & \left(\boldx^{\text{id}}_{ui} = 1\right) \\
        \left[\text{INF} \mathbbm{1}_m^\top, \ldots, \text{INF} \mathbbm{1}_m^\top, \ldots, \text{INF} \mathbbm{1}_m^\top\right] & \left(\boldx^{\text{id}}_{u} = \bold0_m\right)
    \end{cases}.
\end{align}
The update function of the $L_n + n + 1$-th layer is \begin{align} \label{eq: main_mid_first_upd}
    &f_{\text{upd}, \theta_n}^{(L_n + 1)}\left(\left[\hat{\bar{s}}(\boldz_v), \boldx^{\text{id}\top}_v, \boldd_v^{(L_n + n)\top}\right]^\top, \bolda_v^{(L_n + n + 1)}\right) \stackrel{\text{def}}{=} \left[\hat{\bar{s}}(\boldz_v), \boldx^{\text{id}\top}_v, \boldd_v^{(L_n + n)\top}, \boldD_v^{(L_n + n + 1)\top}\right]^\top \in \mathbb{R}^{2m + m^2 + 1}, \\
    &\boldD_v^{(L_n + n + 1)} \stackrel{\text{def}}{=} \min \left(\boldU_v, \bolda_v^{(L_n + 1)}\right) \in \mathbb{R}^{m^2}, \\
    &\boldU_v \stackrel{\text{def}}{=} \begin{cases} \left[\text{INF} \mathbbm{1}_m^\top, \ldots, \text{INF} \mathbbm{1}_m^\top, \overbrace{\boldd_v^{(L_n + n)\top}}^{i\text{th}}, \text{INF} \mathbbm{1}_m^\top, \ldots, \text{INF} \mathbbm{1}_m^\top\right] & \left(\boldx^{\text{id}}_{vi} = 1\right) \\
        \left[\text{INF} \mathbbm{1}_m^\top, \ldots, \text{INF} \mathbbm{1}_m^\top, \ldots, \text{INF} \mathbbm{1}_m^\top\right] & \left(\boldx^{\text{id}}_{u} = \bold0_m\right)
    \end{cases}.
\end{align}
The aggregation function of the $L_n + n + i$-th layer $(2 \le i \le n)$ is \begin{align} \label{eq: main_mid_mid_agg}
    f_{\text{agg}, \theta_n}^{(L_n + n + i)}(\lbb [\hat{\bar{s}}(\boldz_u), \boldx^{\text{id}\top}_u, \boldd_u^{(L_n + n)\top}, \boldD_u^{(L_n + n + i - 1)\top}]^\top \mid u \in \mathcal{N}^-(v) \rbb) &\stackrel{\text{def}}{=} \min_{u \in \mathcal{N}^-(v)} \boldD_u^{(L_n + n + i - 1)},
\end{align}
and the update function of the $L_n + n + i$-th layer $(2 \le i \le n - 1)$ is \begin{align} \label{eq: main_mid_mid_upd}
    f_{\text{upd}, \theta_n}^{(L_n + n + i)}\left(\left[\hat{\bar{s}}(\boldz_v), \boldx^{\text{id}\top}_v, \boldd_v^{(L_n + n)\top}, \boldD_v^{(L_n + n + i - 1)\top}\right]^\top, \bolda_v^{(L_n + n + i)}\right) &\stackrel{\text{def}}{=} \left[\hat{\bar{s}}(\boldz_v), \boldx^{\text{id}\top}_v, \boldd_v^{(L_n + n)\top}, \boldD_v^{(L_n + n + i)\top}\right]^\top, \\
    \boldD_v^{(L_n + n + i)} &\stackrel{\text{def}}{=} \min \left(\boldD_v^{(L_n + n + i - 1)}, \bolda_v^{(L_n + n + i)}\right).
\end{align}
The last update function is defined as follows. \begin{align} \label{eq: main_mid_last_upd}
    f_{\text{upd}, \theta_n}^{(L_n + 2n)}\left(\left[\hat{\bar{s}}(\boldz_v), \boldx^{\text{id}\top}_v, \boldd_v^{(L_n + n)\top}, \boldD_v^{(L_n + 2n - 1)\top}\right]^\top, \bolda_v^{(L_n + 2n)}\right) &\stackrel{\text{def}}{=} \begin{cases}
        \text{MDS}\left(\boldD_v^{(L_n + 2n)}\right)_i & \left(\boldx^{\text{id}}_{vi} = 1\right) \\
        \text{MDS}\left(\boldD_v^{(L_n + 2n)}\right)_{k_v} & \left(\boldx^{\text{id}}_{v} = \bold0_m\right) \\
    \end{cases}, \\
    \boldD_v^{(L_n + 2n)} &\stackrel{\text{def}}{=} \min \left(\boldD_v^{(L_n + 2n - 1)}, \bolda_v^{(L_n + n + i)}\right), \\
    k(v) &\stackrel{\text{def}}{=} \argmin_i {\boldd_v^{(L_n + n)}}_i.
\end{align}
As the diameter of $G_n$ is at most $n$, the propagation is complete after $n$ iterations. Therefore, ${\boldD_v^{(L_n + 2n)}}_{in + j}$ is the shortest-path distance from $u_i$ to $u_j$ with the length of edge $(s, t)$ being $g_n \hat{\bar{s}}(\boldz_s)$. $\text{MDS}\colon \mathbb{R}^{m^2} \to \mathbb{R}^{m \times d}$ runs the multidimensional scaling. Note that in the $(L_n + 2n)$-th layer, each node has the distance matrix $\boldD^{(L_n + 2n)}$ in its embedding, therefore MDS can be run in each node in a parallel manner. If $v$ is in $\mathcal{U}$, $f_{\text{upd}, \theta_n}^{(L_n + 2n)}$ outputs the coordinate of $v$ recovered by MDS. If $v$ is not in $\mathcal{U}$, $f_{\text{upd}, \theta_n}^{(L_n + 2n)}$ outputs the coordinate of the closest selected node, i.e., $u_{k(v)}$.

We analyze the approximation error of the above processes. Let $\boldZ_{\mathcal{U}} = [\boldz_{u_1}, \ldots, \boldz_{u_m}]^\top \in \mathbb{R}^{m \times d}$ be the true hidden embeddings of the selected nodes. By Corollary 4.2 of \cite{sibson1979studies}, the noise $C$ to the distance matrix causes $O(C^4)$ of misalignment of the coordinates. Therefore, if \begin{align}
    \forall u_i, u_j \in \mathcal{U}, \left|{\boldd_{u_i}^{(L_n + n)}}_j - \|\boldz_{u_i} - \boldz_{u_j}\|\right| < \varepsilon' \label{eq: distance_of_U_is_precise}
\end{align} 
holds for some $\varepsilon' > 0$, then \begin{align} \label{eq: MDS_precise}
d_G\left(\text{MDS}\left(\boldD_v^{(L_n + 2n)}\right), \boldZ_{\mathcal{U}}\right) < \frac{\varepsilon^2}{9m}.
\end{align} Let $$\boldP_{\mathcal{U}} \stackrel{\text{def}}{=} \argmin_{\boldP \in \mathbb{R}^{d \times d}, \boldP^\top \boldP = I_d} \left\|\boldC_m \text{MDS}\left(\boldD_v^{(L_n + 2n)}\right) - \boldC_m \boldZ_{\mathcal{U}} \boldP\right\|_F^2.$$ If Eq. \eqref{eq: MDS_precise} holds, for all $i = 1, \ldots, m$, \begin{align} \label{eq: coodinate_of_u_is_precise}
\left\|\left(\boldC_m \text{MDS}\left(\boldD_v^{(L_n + 2n)}\right)\right)_i - \left(\boldC_m \boldZ_{\mathcal{U}} \boldP_{\mathcal{U}}\right)_i\right\|_2 \le \frac{\varepsilon}{3}.
\end{align}

If $n$ is sufficiently large, \begin{align} \label{eq: prob_distance_is_precise}
\text{Pr}\left[\forall v \in V, u_i \in \mathcal{U}, \left|{\boldd_v^{(L_n + n)}}_i - \|\boldz_v - \boldz_{u_i}\|\right| < \min\left(\varepsilon', \frac{\varepsilon}{6}\right)\right] \ge 1 - \frac{\delta}{2}
\end{align} holds from Theorem S.4.5 of \cite{hashimoto2015metric}. Note that although Theorem S.4.5 of \cite{hashimoto2015metric} uses the true $\pi_{G_n}$ while we use $\bolda_v^{(L_n)}$, from Eq. \eqref{eq: scale_num_layer} and \eqref{eq: scale_GNN_invariant}, $|\bolda_v^{(L_n)} - n \pi_{G_n}| \xrightarrow{n \to \infty} 0$ holds surely, and therefore, the theorem holds because the mismatch diminishes as $n \to \infty$.

We suppose the following event $P$: \begin{align}
    &\mathcal{U} \text{ forms an } \frac{\varepsilon}{3} \text{ covering and} \label{eq: U_is_covering} \\
    &\forall v \in V, u_i \in \mathcal{U}, \left|{\boldd_v^{(L_n + n)}}_i - \|\boldz_v - \boldz_{u_i}\|\right| < \min\left(\varepsilon', \frac{\varepsilon}{6}\right). \label{eq: distance_is_precise}
\end{align}
The probability of this event is at least $1 - \delta$ by Eq. \eqref{eq: prob_U_is_covering} and \eqref{eq: prob_distance_is_precise}. Under this event, for all $i = 1, \ldots, m$, \begin{align}
    &\left\|\left(\hat{\boldZ} - \frac{1}{m} \mathbbm{1}_n \mathbbm{1}_m^\top \text{MDS}\left(\boldD_v^{(L_n + 2n)}\right)\right)_{u_i} - \left(\left(\boldZ - \frac{1}{m} \mathbbm{1}_n \mathbbm{1}_m^\top \boldZ_{\mathcal{U}}\right) \boldP_{\mathcal{U}}\right)_i\right\|_2 \\
    &=\left\|\left(\boldC_m \text{MDS}\left(\boldD_v^{(L_n + 2n)}\right)\right)_i - \left(\boldC_m \boldZ_{\mathcal{U}} \boldP_{\mathcal{U}}\right)_i\right\|_2 \\
    &< \frac{\varepsilon}{3}. \label{eq: MDS_is_precise}
\end{align}
holds by Eq. \eqref{eq: distance_is_precise}, \eqref{eq: distance_of_U_is_precise}, \eqref{eq: MDS_precise}, and \eqref{eq: coodinate_of_u_is_precise}, and the definition of $\hat{\boldZ}$ (Eq. \eqref{eq: main_mid_last_upd}) and $\boldC_m$. Under event $P$, for any $v \not \in \mathcal{U}$, there exists $u_i \in \mathcal{U}$ such that $\|\boldz_{v} - \boldz_{u_i}\| \le \frac{\varepsilon}{3}$ by Eq. \eqref{eq: U_is_covering}. By applying Eq. \eqref{eq: distance_is_precise} twice, \begin{align}
    {\boldd_v^{(L_n + n)}}_{k(v)} = \min_i {\boldd_v^{(L_n + n)}}_i &< \frac{\varepsilon}{2}, \\
    \left\|\boldz_{v} - \boldz_{u_{k(v)}}\right\| &< \frac{2\varepsilon}{3}. \label{eq: z_v_and_z_k(v)_is_colose}
\end{align}
Then, \begin{align}
    &\left\|\left(\hat{\boldZ} - \frac{1}{m} \mathbbm{1}_n \mathbbm{1}_m^\top \text{MDS}\left(\boldD_v^{(L_n + 2n)}\right)\right)_v - \left(\left(\boldZ - \frac{1}{m} \mathbbm{1}_n \mathbbm{1}_m^\top \boldZ_{\mathcal{U}}\right) \boldP_{\mathcal{U}}\right)_v\right\|_2 \\
    &\stackrel{\text{(a)}}{=} \left\|\left(\boldC_m \text{MDS}\left(\boldD_v^{(L_n + 2n)}\right)\right)_{k(v)} - \left(\left(\boldZ - \frac{1}{m} \mathbbm{1}_n \mathbbm{1}_m^\top \boldZ_{\mathcal{U}}\right) \boldP_{\mathcal{U}}\right)_v\right\|_2 \\
    &\stackrel{\text{(b)}}{\le} \left\|\left(\boldC_m \text{MDS}\left(\boldD_v^{(L_n + 2n)}\right)\right)_{k(v)} - \left(\left(\boldZ - \frac{1}{m} \mathbbm{1}_n \mathbbm{1}_m^\top \boldZ_{\mathcal{U}}\right) \boldP_{\mathcal{U}}\right)_{k(v)}\right\|_2 \\ & \quad + \left\|\left(\left(\boldZ - \frac{1}{m} \mathbbm{1}_n \mathbbm{1}_m^\top \boldZ_{\mathcal{U}}\right) \boldP_{\mathcal{U}}\right)_{v} - \left(\left(\boldZ - \frac{1}{m} \mathbbm{1}_n \mathbbm{1}_m^\top \boldZ_{\mathcal{U}}\right) \boldP_{\mathcal{U}}\right)_{k(v)}\right\|_2 \\
    &\stackrel{\text{(c)}}{=} \left\|\left(\boldC_m \text{MDS}\left(\boldD_v^{(L_n + 2n)}\right)\right)_{k(v)} - \left(\left(\boldZ - \frac{1}{m} \mathbbm{1}_n \mathbbm{1}_m^\top \boldZ_{\mathcal{U}}\right) \boldP_{\mathcal{U}}\right)_{k(v)}\right\|_2 + \|\boldz_v - \boldz_{k(v)}\|_2 \\
    &\stackrel{\text{(d)}}{<} \left\|\left(\boldC_m \text{MDS}\left(\boldD_v^{(L_n + 2n)}\right)\right)_{k(v)} - \left(\left(\boldZ - \frac{1}{m} \mathbbm{1}_n \mathbbm{1}_m^\top \boldZ_{\mathcal{U}}\right) \boldP_{\mathcal{U}}\right)_{k(v)}\right\|_2 + \frac{2}{3} \varepsilon \\
    &\stackrel{\text{(e)}}{<} \varepsilon, \label{eq: z_is_precise_for_non_selected}
\end{align}
where (a) follows because $\hat{\boldZ}_v = \text{MDS}\left(\boldD_v^{(L_n + 2n)}\right)_{k(v)}$ by Eq. \eqref{eq: main_mid_last_upd} and by the definition of $\boldC_m$, (b) follows by the triangle inequality, (c) follows because $\boldP_{\mathcal{U}}$ is an orthogonal matrix, (d) follows by Eq. \eqref{eq: z_v_and_z_k(v)_is_colose}, and (e) follows by \eqref{eq: MDS_is_precise}. By Eq. \eqref{eq: MDS_is_precise} and \eqref{eq: z_is_precise_for_non_selected}, the the distance between the true embedding $\boldz_v$ and the estimate $\hat{\boldz}_v$ is less than $\varepsilon$ with rigid transformation $\boldP_{\mathcal{U}}$ and translation $- \frac{1}{m} \mathbbm{1}_n \mathbbm{1}_m^\top \text{MDS}\left(\boldD_v^{(L_n + 2n)}\right)$. As these transformation parameters are suboptimal for Eq. \eqref{eq: d_G}, these distances are overestimated. Therefore, Under event $P$, $d_G(\hat{\boldZ}, \boldZ) < \varepsilon$, which holds with probability at least $1 - \delta$.

\end{prooftn}

The definitions of the layers prove Theorem \ref{thm: main-bound}.

\section{Technical Remarks}

\textbf{Remark (Global Information).} The existing analyses of the over-smoothing effect \cite{li2018deeper, oono2020graph} show that GNNs with too many layers fail. Therefore, GNNs cannot have wide receptive fields, and GNNs cannot aggregate global information. By contrast, our analysis shows that GNNs can obtain global information, i.e., $\boldz_v$. This result provides a new insight into the understanding of GNNs. Note that the assumptions of the existing analyses \cite{li2018deeper, oono2020graph} do not hold for our GNN architectures. Therefore, our results do not contradict with the existing results.

\textbf{Remark (Positions as Explicit Node Features are Redundant).} In some applications, the graph is constructed from observed features, and $\{\boldz_v\}$ are available as the explicit node features \cite{han2019gcn, wang2020am, han2022vision}. For example, each node represents a position of interest in spatial data, the graph is constructed by nearest neighbors based on geographic positions, and the positions are included in the explicit node features $\boldx_v$. Our main results show that such position features are asymptotically redundant because GNNs can recover them solely from the graph structure. In practice with finite samples, the position features can be informative, and they can introduce a good inductive bias, though.

\textbf{Limitation (High Node Degree).} We assume high node degrees in Assumption 4 and in the experiments, i.e., $d_v = \omega(n^{\frac{2}{d + 2}} \log^{\frac{d}{d+2}} n)$. Note that $d_v = \Theta(\log n)$ is required for random graphs to be connected \cite{erdos1959random}, so we cannot reduce node degrees so much from a technical point of view. Having said that, there is indeed room for improvement by a factor of $\left(\frac{n}{\log n}\right)^{\frac{2}{d + 2}}$, which can be indeed large when $d$ is small. This bound is common with \cite{hashimoto2015metric}, and improving the bound is important future work.

\textbf{Remark (Dimensionality of the True Features).} We need to specify the number of dimensions of the true features, which are not necessarily available in practice. Specifying higher number of dimensions than the true one is not so problematic, as the lower dimensional features are recovered in the subspace of the entire space. In practice, we can find a good dimensionality by measuring a reconstruction loss in an unsupervised manner. Namely, after we recover the features, we construct a nearest neighbor graph from it. If it does not resemble the input graph, the dimensionality may not be sufficient.


\end{document}